# PREDICTING SURFACE TEXTURE IN STEEL MANUFACTURING AT SPEED

MAKING THE "EXCEPTIONALLY FAST" ROCKET MODEL FASTER

ALEXANDER J. M. MILNE

Department of Computer Science, Swansea University, Swansea, UK

XIANGHUA XIE

Department of Computer Science, Swansea University, Swansea, UK

Control of the surface texture of steel strip during the galvanizing and temper rolling processes is essential to satisfy customer requirements and is conventionally measured post-production using a stylus. In-production laser reflection measurement is less consistent than physical measurement but enables real time adjustment of processing parameters to optimize product surface characteristics. We propose the use of machine learning to improve accuracy of the transformation from inline laser reflection measurements to a prediction of surface properties. In addition to accuracy, model evaluation speed is important for fast feedback control. The ROCKET model is one of the fastest state of the art models, however it can be sped up by utilizing a GPU. Our contribution is to implement the model in PyTorch for fast GPU kernel transforms and provide a soft version of the Proportion of Positive Values (PPV) nonlinear pooling function, allowing gradient flow. We perform timing and performance experiments comparing the implementations.

## 1 INTRODUCTION

In steel strip manufacturing, monitoring the surface texture resulting from the galvanizing and temper rolling process inline (i.e. during production) allows for fast feedback for line parameter adjustments. Conventional slow measurements performed during post-production using a stylus are essential for ensuring the strip steel produced has the correct properties and satisfies customer requirements. However, post-production measurements give slow feedback which is only useful for later rolls. Therefore, fast feedback control is required for real time changes to steel surface. This has the potential to reduce value loss by reducing the likelihood of the surface texture being found inadequate during postproduction testing. However, for adjustments to be made based on fast feedback, it is essential that the inline measurements are accurate. Compared to conventional stylus methods of post-production measurements, the method of inline measurement we are using are not as consistent.

The method for inline measurements we use involves firing a laser at the steel surface and measuring the reflected angle. These angles represent the gradient of the steel in the measurement positions and can be used to calculate the surface profile and therefore, the surface properties. However, as previously mentioned, this method produces inconsistent results. The exact reason is unknown, though we speculate that it may be because the sensors measuring the angle are discrete, or because they are lined up in one dimension. The steel surface can reflect light in two dimensions; thus, the array might not capture sufficient information for stitching the gradients when the light reflects away from the measurement dimension. Another potential reason is that the data does not have a high enough resolution in the time dimension, with not enough sampling. To overcome these issues, we propose the use of machine learning to learn the unexplained transformation from our raw reflected light intensity measurements to surface properties, thus allowing us to make use of fast feedback control. We demonstrate the effectiveness of this approach by predicting the Ra roughness parameter (defined in section 3), however the same method can be applied for predicting many other surface parameters, either together in the same model, or separately.

For the use in fast feedback control, accuracy is not the only consideration when modelling the transformation; the speed of the model chosen is also of great importance. Each sample from the dataset used in this study is relatively large. It has a high resolution with 5cm of sample reading having 20 channels and 50,000 timesteps. The line speed is typically between 50cm/s and 300cm/s, meaning that to perform prediction at line speed we need to be able to run 60 examples per second. The combination of high resolution and fast line speed compounds the need for a faster model for our use case.

We have chosen to interpret this problem as a Time Series Extrinsic Regression (TSER) problem due to the possibility for each of the feature dimensions having independent partial occlusion, as each is provided by a separate sensor. However, this problem could also be interpreted as an image regression problem due to the spatial relation in two dimensions. The decision was made in hope of a more robust model, able to deal with variation between sensors. More is written about this in the data normalization section below.

Time series classification (TSC) problems are related to TSER problems [3]. Both TSC and TSER take input data and transform it into a representation which is used by an output model or layer to predict either a class, in the case of TSC, or continuous value, in the case of TSER. Unlike in time series forecasting, where the predicted value is a future value of the series or a value of a different data type which is more dependent on recent values, TSER is a more general form where the output data can take any type and be related to any part of the input series, such as in our use case.

The ROCKET model performs well while being exceptionally fast on univariate TSC [1]. It has also been shown to work on TSER problems [3]. It has both a univariate and multivariate version, with success on multivariate problems documented [10]. For its success when applied to both multivariate problems and TSER, the ROCKET model is a good choice for our use case.

Computation speed is an important aspect when producing a model. With increased speed and efficiency, it becomes feasible to create more complicated models with more calculations. The highly influential "AlexNet" [11], can attribute a lot of its success to the GPU speed up. The paper showed that the model beat other state of the art models by a significant margin on the ImageNet dataset [12], showing that increased depth and complexity of the model improves performance. This was made feasible through GPU computation.

In this paper we have chosen to reimplement ROCKET on GPU for the purpose of performing predictions at a pace more in line with the steel production line speed of our use case. However, the benefits of the GPU implementation go beyond our use case, as ROCKET is a well-established baseline, so improving its speed offers a better baseline for comparisons; also, given its increased speed, further improvements to its accuracy could be made by increasing the default number of kernels. Therefore, this work has the possibility for use in the wider TSC and TSER communities, with the potential to speed up the model generally or by making adaptations feasible to increase model complexity in ways that might improve performance in future work.

We also show that the GPU implementation is faster than the CPU-based MINIROCKET model [13] which was later introduced as an improvement to the ROCKET model in terms of computational efficiency while maintaining similar accuracy.

Our main contributions in this paper are as follows:

- We provide two implementations of the Proportion of Positive Values (PPV) nonlinear pooling function. The first version is achieved by decomposing PPV into the Heaviside step function and a mean. For the second, we propose the use of a soft version PPV, which in our experiments does not negatively impact performance when its parameter is greater than 3 and has a very minor improvement with our data. The soft version is achieved by approximating the Heaviside step function using a sigmoid function and a mean. This version has the added benefit of allowing the potential for gradient flow and training, but we do not make use of this in the paper.
- Employing the use of GPU, we implement the ROCKET transformation in the PyTorch [20] library, yielding a significant speed up with our data. Using the best speeds for both versions on our data, the speed goes from a CPU result of 0.65 transformations per second to 5.52 transformations per second, a 749% speed increase.
- We show that the model performs well for our use case and produces Ra values from the laser readings within the standard deviation of the corresponding stylus measurements.
- We explore the effect on speed of different batch sizes comparing our GPU implementation against the CPU sktime [17,18] implementation and provide results in the form of timing experiments.
- We also explore the effect on performance of different hyperparameters used with the softPPV function.
- We provide all code in the GitHub repository [19].

The rest of the paper is structured as follows. Section 2 discusses the related work. Section 3 describes the dataset used for our experiments. Section 4 gives our methodology. Section 5 discusses our experiments and results. Section 6 presents our conclusions.

## 2 RELATED WORK

Elangovan et al. [2] propose the use of machine learning for characterizing metal surface roughness Ra in a different context to our use case, instead characterizing the Ra of metal objects turned on a lathe. They use multiple regression analysis on statistical features extracted from the vibration signal from the tool, in addition to machining parameters such as tool wear, speed, etc. The extracted statistical features were simple: mean, skewness, etc. Better features might be extracted by applying kernels to the signal.

Machine learning is commonly applied in manufacturing to perform automated visual inspection of the surface [14,15,16]. This can be used in strip steel production for detection of defects. These techniques typically use an image sensor or similar and take repeated pictures of the surface during production. They perform machine vision techniques on this image data in order to locate or classify defects on the surface. However, these techniques are not applicable to our use case, as we are not interested in finding and locating defects, but characterising the surface of the steel using roughness Ra.

Our problem is interpreted as a TSER problem, but there is more research attention in the field of Time Series Classification (TSC). As TSC is a more mature problem, in the paper by Chang Wei Tan et al. [3], the authors adapt top TSC models for use in TSER problems and apply them to 19 datasets. The ROCKET model performs the best, followed by InceptionTime, FCN, XGBoost, ResNet, etc. They note that although ROCKET performs well, it does not produce results significantly distinguishable from classical algorithms such as XGBoost and Random Forest, which ignore temporal ordering of the data. In TSC problems, algorithms ignoring temporal information perform significantly worse than those that do not, suggesting more research is needed specific to TSER problems.

ROCKET, the best performer from the experiments by from Chang Wei Tan et al. [3], was proposed by Angus Dempster et al. [1] as "exceptionally fast and accurate", in contrast, most other state of the art models are computationally expensive. This model applies 10,000 kernels each with random parameterisation, followed by two global pooling operations for each kernel, producing 20,000 features used by a ridge classifier for predictions. The pooling operations are a max pooling operation and a proportion of positive values operation (PPV). The model was originally proposed as a TSC model, but the ridge classifier can be replaced by a ridge regressor as in [3]. The ROCKET paper results show that it beats TS-Chief [8], HIVE-COTE [7], and InceptionTime [4] as well as other models in terms of accuracy, while also having the shortest training phase when applied to the 85 UCR TSC datasets [6].

Angus Dempster et al. later proposed MINIROCKET [13] with a focus on improving efficiency compared to the earlier proposed ROCKET model. MINIROCKET represents a significant advance in accuracy relative to computational cost. They gain this speed improvement by performing convolution operations via addition, which is possible due to their utilisation of binary kernels. They also make some changes to the random kernel parameterisation by using a set of predetermined weights with biases samples from the input data. They also remove the max pooling feature, only doing the PPV pooling operation. They find that MINIROCKET is 75 times faster on larger datasets and with relatively similar accuracy when compared to ROCKET, a model which is already fast compared to other models in the literature.

Later, Chang Wei Tan et al. uses the computational efficiency improvement gained from the MINIROCKET model to increase model complexity in MultiRocket [21]. The model performs a first differences transform to the input data and feeds this along with the original input data to the same kernel transformation used in MINIROCKET. They also add three additional pooling operations to the model along with the PPV pooling to total four features per kernel. This results in ~50,000 features compared to the ~10,000 of MINIROCKET. This model produces worse results on our data compared to the other variants so it not useful for our problem. However, it givens an example of how our work improving the ROCKET model's computational performance can be used by others to increase model complexity for better results.

In [4], authors Fawaz et al. propose the InceptionTime model for TSC, inspired by the Inception-v4 architecture [5]. The deep learning model's results in terms of accuracy are on par with HIVE-COTE while being significantly more computationally efficient. The model consists of an ensemble of five differently initialised Inception networks. The networks have two Inception blocks with residual connections adding each block's input to its output. The blocks have 3 Inception modules, each with 4 output paths. One to three are generated by applying a bottleneck layer to the input, reducing the number of dimensions, followed by 10, 20, and 40 length convolutional filters applied to the output of the bottleneck layer. The fourth is a max pooling layer applied to the input followed by a bottleneck layer. Finally, there is a Global Average Pooling (GAP) layer followed by a softmax classifier layer with an equal number of neurons to number of classes.

Alejandro Ruiz et al. [10] analyse TSC models on multivariate datasets, as opposed to the usual univariate UCR archive date [6] used in comparisons. The analysis shows that HIVE-COTE (with components all built independently on each dimension), CIF, ROCKET and InceptionTime are significantly better than the baseline DTW model. They find InceptionTime to be the top performing algorithm in terms of counting the problems won and say that it should be the starting point for future work with neural networks on these problems. However, they find that the ROCKET model has the lowest average rank while being by far the fastest classifier. The authors claim it to be the clear winner of their experimental study and that it should be used as a benchmark for any new models in the domain.

## 3 THE METHODOLOGY

Our methodology uses a ROCKET model adapted for regression, which we have reimplemented for improved computational efficiency. The model consists of three main parts: (1) Random convolutional transformation: here, convolutional kernels that have been initialized with random parameters and weights are applied to the input data, performing a linear transformation. (2) Pooling operation features: here, two nonlinear pooling operations are applied to each of the kernel outputs from the previous step, resulting in two features per kernel per example. (3) Linear regressor: here, the linear regressor takes the features from the pooling operations and uses them to make predictions. In the original ROCKET paper, a linear classifier is used, but success has also been shown using a regressor with the random kernel features [3]. We leave the random convolutional transform the same as in the original, simply implementing it for use on the GPU. We do however make changes to the pooling operations by changing the PPV function to be more suitable.

The ROCKET model design allows it to be trained exceptionally fast since the random feature transform is not trained and is performed only once, leaving only the linear regressor to train. However, for our use case, it is not fast enough at making predictions since only the linear regressor is faster during inference compared to training; the random kernel transform takes the same time whether performing training or evaluation.

The original implementation of the ROCKET model forgoes GPU, performing the random kernel transform on the CPU. This leaves a potentially substantial speedup for the random feature transform phase of the model. Our implementation utilizes the PyTorch [20] library to perform the random kernel transform and the nonlinear pooling PPV operation. The output from this model is then fed into a ridge regressor, however, a neural network based linear regressor could also be used to have an end-to-end PyTorch model on the GPU. In this section we give an overview of the ROCKET model's random features and the changes we have made for the GPU implementation. Our code can be found in our git repository [19].

### 3.1 The Random Kernel Initialization

As previously mentioned, we have decided to leave the random kernels' initialization parameters as specified by the original authors of the ROCKET model and have not investigated changing the parameters to suit our data. We have done this so that we can perform fair timing experiments for our comparison later, as the speed improvement is the aspect of this paper most useful for the reader. One change we did make was to edit the functions to use 32bit precision floating point operations, as we are using 32bit in our PyTorch implementation.

For cross comparison, we generate the kernels using the sktime implementation edited to use 32bit and load these kernels into the PyTorch model. Therefore, as in [1], we have 10,000 1d convolution operations, each with random kernels with:

- Weights selected from a normal distribution between 0 and 1. Each weight is mean centered.
- Biases from a uniform distribution between -1 and 1.
- A kernel size selected uniformly from 7, 9, and 11.
- A random choice between padding to produce the same sized output as input, or no padding.
- A random dilation sampled from an exponential scale up to input length.

- A random selection of which channels will be convolved.

## 3.2 Proportion of Positive Values (PPV) and Soft Proportion of Positive Values (softPPV)

This part of the model has seen changes with different two implementations resulting in different efficiency results, while maintaining accuracy, as shown later in the experiments section. The Proportion of Positive Values (PPV) function is a pooling function which takes a vector z, counts the occurrences of positive values, and divides by the number of elements in z. This can also be expressed as applying a Heaviside step function, Equation 1, and then taking the mean, Equation 2.

$$H(x) := \begin{cases} 1, & x > 0 \\ 0, & x \leq 0 \end{cases} \quad (1)$$

$$PPV = \frac{1}{n}\sum_{i=1}^{n} H(z_i) \quad (2)$$

Since we are implementing the layer in PyTorch which is commonly used to implement deep learning models, we have decided to include a soft version of PPV. To soften this layer of the model, we use an approximation of the Heaviside step function, Equation 3. This has the benefit of making it differentiable, giving it the potential for gradient flow. The approximation used is a shifted sigmoid function that produces values of one when the input is large and values of zero when the input is small. We would like to point out there this layer might introduce the same problems as when using sigmoid layers as nonlinearities, where gradients become small when the inputs are either large or small.

$$H_{soft}(x) = \frac{1}{1+e^{-\lambda x - 3}} \quad (3)$$

$$SoftPPV = \frac{1}{n}\sum_{i=1}^{n} H_{soft}(z_i) \quad (4)$$

The λ parameter of this function changes the "softness" and as $\lambda \to \infty$ the function approaches the Heaviside step function. The -3 term is included to shift the function so that negative values are closer to zero than positive values are to one, reducing the effect of negative values on the mean. The -3 term has a diminished effect with large λ, as seen in Figure 1.

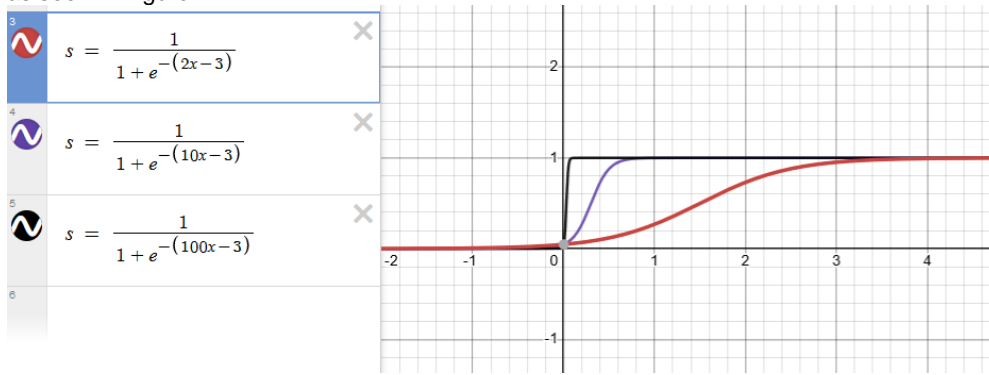

Figure 1: The effect if varying λ in the soft Heaviside function. λ values shown in the plot from softest to hardest approximations are λ =2, λ =10, λ =100.

### 3.3 Regressor

We have chosen to use a simple ridge regressor for our predictions. The 10,000 random kernels are applied to each input example, followed by the two global pooling operations to produce two features per kernel or 20,000 features per example. These features are used by the ridge regressor to make predictions.

## 4 THE DATA

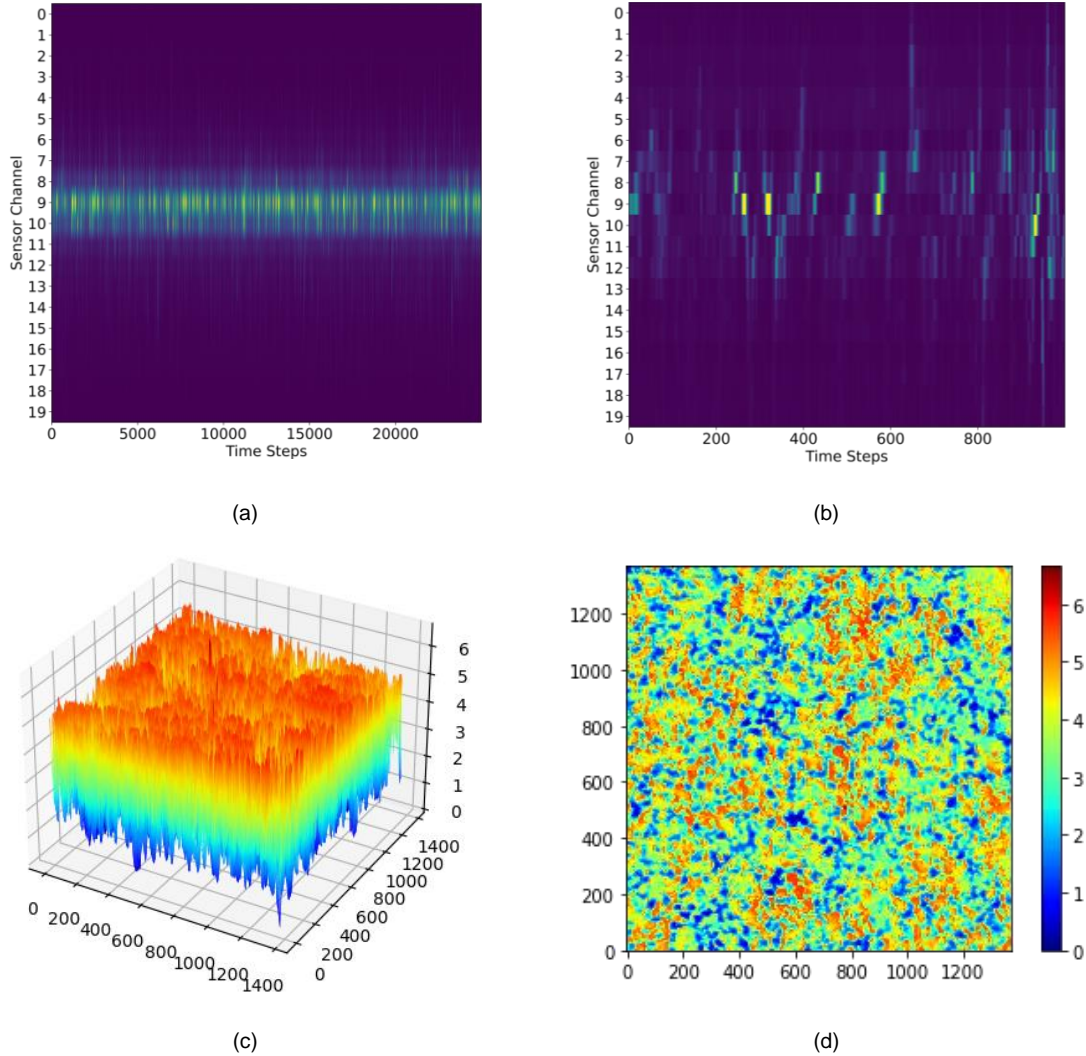

Figure 2: (a) and (b) Show a section of an example from the input data. It is an intensity plot containing intensities across the 20 sensors measuring the laser reflection data through time as the steel moves through the measurement device. (c) and (d) Show a confocal microscope images taken of a section of steel surface with measurements all being in μm.

In this paper we perform experiments on the raw data provided by our industrial partner in order to judge the feasibility of our approach in terms of both speed and accuracy. From the raw data we have created a supervised learning dataset consisting of input features and labels. It consists of measurements taken from 49 samples of different steel products. From each sample multiple measurements were taken in different positions. The test set is selected by randomly choosing a measurement example from each of the steel samples. This results in 726 examples split into 677 training examples and 49 testing examples.

### 4.1 Input Features

The input data are generated by firing a laser pulse at the surface of the steel and capturing the reflected light intensities across a semicircular one-dimensional array of sensors. This is repeated for the chosen number of timesteps as the steel continuously moves through the measurement device at a constant rate. The measurement device aims to capture the gradient of the surface of the steel using the reflected angles, which can then be transformed into a line profile with surface properties calculated from this. However, unknown properties of the system make the relationship between this data and the line profile unknown. A raw input example is shown in Figure 2.

Each of the 726 samples has 20 input channels, each corresponding to one of the light sensors, and 50,000 timesteps with a distance of 0.8um between each measurement, in total capturing 4cm of steel. The values captured by the sensors are integer light intensities between 0 and 255. A different number of laser measurements were taken for each steel sample, resulting in some samples having significantly more input data. A larger proportion of the samples (24 of them) have 10 measurements, while 13 samples have 25 measurements. The remaining number of samples to number of measurements are 4, 3, 2, 1 with 5, 26, 24, 23 measurements respectively. The reason for different number of measurements for each sample in the data set is unknown.

### 4.2 Labels

Ideally, the label for each sample should be the Ra roughness parameter from the stylus measurement of the track which corresponds with the input data. However, it is not feasible to gather this data as we do not have the capability to run the stylus in the same track. Even if this were possible, the width of the laser and stylus are not the same. As a result, we must use the mean Ra measurement from all stylus measurements on the steel sample.

Each sample has 21 stylus measurements creating line profiles. The line profiles are transformed into roughness profiles by removing the longer wavelengths with a high pass filter as in [9]. The Ra parameter is the mean deviation of the roughness profile Z.

$$\text{Ra} = \frac{1}{n}\sum_{i=1}^{n}|z_i|$$

For each steel sample the mean of the 21 Ra statistics is then calculated and this mean value is used as the label. This methodology could be used for other surface parameters, but we have chosen only to predict the Ra as this is the most important for our use case. For our data the minimum Ra value calculated is 0.605 and the maximum is 1.834.

### 4.3 Feature Scaling

For feature scaling we consider two possibilities: normalize each channel independently, or together. Normalizing the channels together seems the obvious choice due to the input data being 2 dimensional, theoretically being image data with the sensors each providing a pixel in the width dimension and the measurement device scanning a certain distance to create the height dimension. Normalizing each channel independently removes information in the data regarding the relationship between different channel intensities, however we know that this information cannot always be trusted. This is because each of the measurement sensors are separate devices, meaning they have noise issues and require different gain values to be applied to the output signals to make them consistent. They can also suffer from different inaccuracies independently from one another caused by, for example, occlusions by dirt or dust. If a sensor becomes partially occluded by residue from the manufacturing process, the data it collects should be similar to data collected if the sensor was not occluded, but with lower intensities.

For these reasons, we have chosen to normalize the input data across each of the 20 sensors independently. Additionally, as we have chosen to regard this problem as a TSER problem, the models we have used treat the channels as independent, so it also makes sense to treat the channels as independent for normalization. Our aim is for the model to learn the relationship without the inter-channel relational intensity information, creating a more robust model. Within this dataset, due to all the measurements being taken with the measurement device in a consistent state, we do not expect the problems relating to changes in noise or occlusion to be present, and as such results might be good using either scheme. During initial experiments it was found that the accuracy of the model was unaffected by using either normalization scheme. However, with the all-together scheme, we would expect worse generalization on new data as the measurement system degrades over time before adjustments. Further work is needed to decide whether during production, or when using a larger dataset, the normalization scheme makes significant difference. To normalize the data, we mean center each channel at zero and divide by its standard deviation.

## 5 EXPERIMENTS

In this section we perform experiments showing speed comparisons of the kernel transform in PyTorch and sktime. Since the authors of the ROCKET model have provided MINIROCKET as a faster alternative to ROCKET with similar accuracy, we also include timing experiments for MINIROCKET. Additionally, we perform MSE comparisons when using different values of $\lambda$ for the softPPV pooling operations to show that using softPPV does not come with an accuracy cost. We have two testbed machines; one has an R5 3600x AMD CPU and an Nvidia 2080ti GPU while the other has an R9 5950x AMD CPU. The sktime ROCKET and MINIROCKET transformation implementations that we use for these experiments run on the CPU and are specified to use all cores. We have made slight alterations to the sktime code such that it makes use of 32bit (FP32) precision instead of the original 64bit (FP64), as 32bit precision is used by the GPU-ROCKET implementation. We also note that there is the potential for further speedup by reducing floating point precision to FP16 compute and also when deploying the PyTorch model using TorchScript, which allows the code to be run in a high-performance environment such as C++. However, experimenting with precision and TorchScript is out of scope for this paper. All experiments are run using random kernels that have been generated using the sktime function prior to testing.

### 5.1 Timing the ROCKET transform: PyTorch (GPU) vs sktime (CPU)

The speed of the transformation is important to our use case, with 60 predictions a second being the requirement to make predictions for the entire steel strip in time with the production line. The experiments are each run sequentially, with no break, in a loop with the ordering being a random permutation of the experiments. The experiments use a batch size between 1 and 1000 for the GPU experiments, and for the CPU experiments, a batch size between 1 and 255. This is because the speed increase by transforming more examples at the same time tapers off much sooner and the experiments take much longer to run. For a batch size of 1 the model transforms a single example, and for a batch size of 1000, 1000 examples are transformed together.

The experiments in Figure 3 show the speed efficiency benefit of performing the transformation on the GPU compared to the CPU, as well as the difference between using the Heaviside implementation of PPV and softPPV for the nonlinear pooling operation. When comparing sktime ROCKET to the PyTorch GPU model, the PyTorch experiments become significantly faster than the sktime experiments as the number of examples in the input is increased. This is likely due to better parallelisation ability and faster floating-point computation on the GPU. The speed increase is better shown in Figure 3b where the results are plotted in terms of the number of examples per second for different input sizes. The slowest experiment is the sktime ROCKET implementation run on the 3600x CPU with a peak speed of 0.22 transformations per second (t/s), beaten by the same model run on the 5950x CPU with a peak of 0.65 t/s. The next fastest is the PyTorch implementation using softPPV, with a peak speed of 5.12 t/s at a batch size of 822. The fasted variant is the transformation incorporating the Heaviside function which has a peak speed of 5.52 t/s at a batch size of 796.

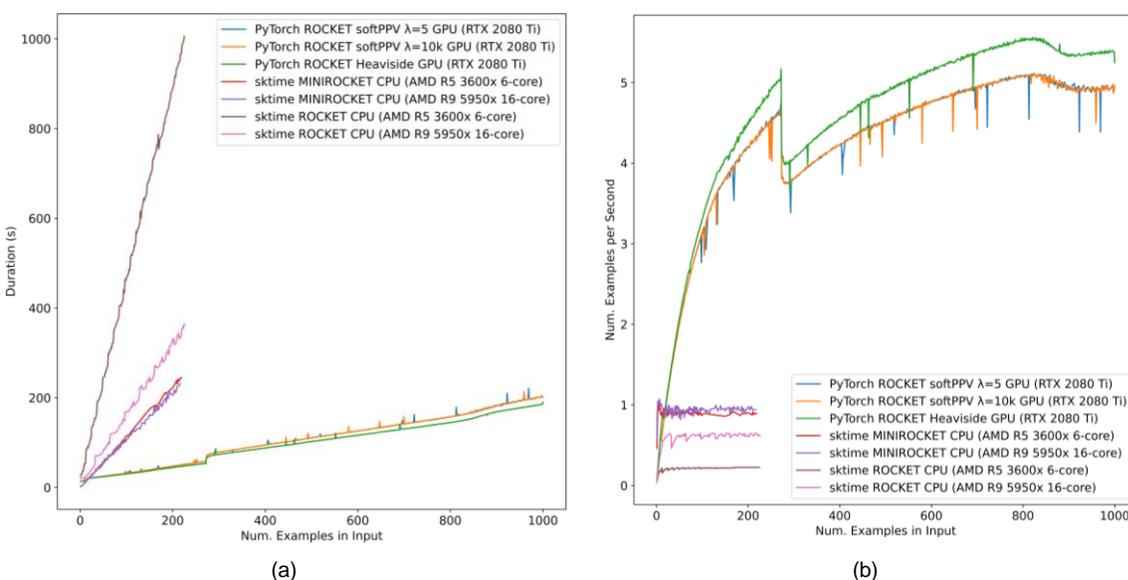

Figure 3. Timing the transform duration for our PyTorch implementation of ROCKET and sktime ROCKET with increasing input sizes and using different testbed systems and different implementations. Figure 3a plots the number of examples in the model's input vs the duration the model took to perform the random kernel transformation. Figure 1b plots the speed (number of predictions per second) of the model

Figure 1b shows that, for all the ROCKET model experiments, similar speeds are achieved for smaller batch sizes, which then diverge as they start to plateau. The plateau is likely due to the maximal capacity for parallelisation being reached. Comparing the two sktime ROCKET experiments run on different CPUs shows that even using the significantly faster CPU, the speedup still plateaus significantly before the GPU result.

The GPU results show a significant drop in performance when stepping up to a batch size of 273. We have eliminated some of the potential reasons for this by performing the experiments non sequentially, such as the potential for it being an overheating problem resulting in thermal throttling. Also, the GPU does not fill all its memory even when performing the experiments with a batch size of 1000. Therefore, the reason for this decrease in performance is unknown. The large spikes in the GPU timings are likely caused by the memory being closed to full and additional time being required to perform memory operations.

The MINIROCKET experiments are faster that the ROCKET experiments when both are using very small batch sizes; however, the GPU ROCKET implementation overtakes sktime MINIROCKET at a batch size of 20 as shown in in Figure 3a. The maximum speed of the MINIROCKET model is 1.01 t/s at a batch size of 3 for the 3600x, and 1.07 t/s at a batch size of 4 for the 5950x. In our experiments the MINIROCKET model doesn't scale as well with CPU core count as the ROCKET model. MINIROCKET results are very similar when comparing the 6 core processor to the significantly more powerful 16 core processor, whereas when comparing the ROCKET model on the two CPU's there is a significant difference.

## 5.2 MSE Ridge Regressor Loss Using Different softPPV parameters

In these experiments we show that there is loss no penalty to using the softPPV approximation compared to the original PPV function for our data. We train a ridge regressor on the sktime ROCKET transformed features, as well as the transformed features from multiple PyTorch implementation versions, each with a different softness parameter affecting the sharpness of the gradient. We then compare the mean squared error (MSE) loss. Training is performed using the transformed training data and testing is performed on the transformed test data withheld during training.

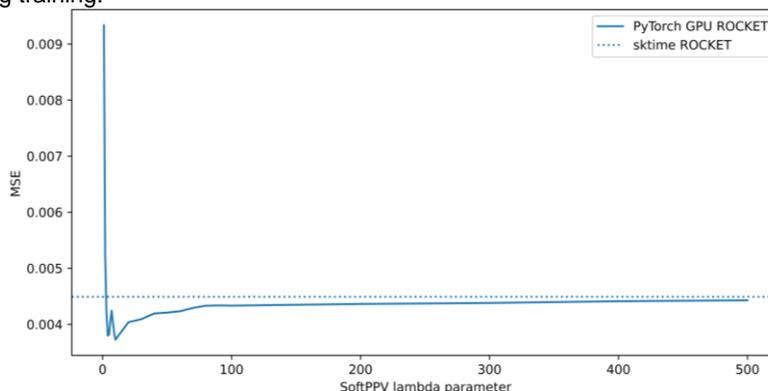

Figure 4. The MSE of ridge regressors trained on the output of the PyTorch implementation of ROCKET with softPPV, given a range of different softPPV hyperparameters.

The testing results in Figure 4 show similar performance for the approaches. For softPPV parameter values of 3 and larger, slightly better results are produced, with larger values trending to a similar MSE as the sktime PPV result. The only times using the softPPV function produced worse results were when the parameter was 1 or 2. Parameter values of 1, 2, 3, and 4 produced MSEs of 0.0093, 0.0052, 0.0042, and 0.0038. The improvement over the sktime PPV implementation is minor; however, it shows that the benefits do not come at a cost to performance when a parameter of 3 or above is chosen. As the softPPV parameter value is increased, the transformation output converges to the same output as PPV, leading to the convergence of MSE results for the two approaches.

### 5.3 Usability of the predictions for our use case

Our experiments show that the model provides results that are useable in production. Using the sktime ROCKET features, the MSE on the test set was 0.0045, and we observed near identical results for the faster GPU implementation. Given that there is a standard deviation in stylus measurements from within the same steel samples of 0.058, the MSE is within this and thus the mean of the results is within measurement error.

### 6 CONCLUSION

In conclusion, we have provided a PyTorch implementation of the ROCKET random feature transform and produced results showing that using this implementation with a GPU is significantly faster. It is 749% faster when comparing the best speeds between sktime and the fastest GPU implementation. We have also made changes to the PPV pooling in order to allow gradient flow such that the pooling operation can be used with trained models. Our MSE results show softPPV doesn't come at a cost to the model results. We also provide results showing the effect on the error when varying the hyperparameter controlling the steepness of the gradient in the softPPV function. Relating to the usability of the model for our use case, we find that the results produced by the model are within the stylus measurement error, and that we are able to perform slightly above 5 transformations a second. This is compared to the 60 laser readings per second taken given the fastest line speed, or 10 per second for the slowest line speed. This means that we have Ra predictions for a large percentage of laser measurements taken, giving line operators fast feedback control.

Future work could focus on using the efficiency gains to experiment with increased model complexity in aid of improved performance. One example of how this might be achieved is shown by the MultiRocket [21] model's changes from MINIROCKET, using the efficiency gains MINIROCKET to improve accuracy while still being very efficient.

One example of how model complexity could be increased is by utilizing additional random kernels above the default, but as shown in the ROCKET this has diminishing results. Other means should be investigated further. Additionally, given that the model is implemented in PyTorch and the softPPV allows for backpropagation, training the weights could be attempted to further improve the model performance. However, this might be less efficient than a deep learning model due to the single layer nature of the transform. Given the success of PPV in the ROCKET model, experiments could be done to judge the viability and accuracy when using softPPV as the global pooling operation as opposed to the normal GAP layer in current deep learning models.

If further improvements in efficiency are required, gains can be made by pruning some percentage of the random kernels based on the feature importance from the ridge regressor, as many of the random filters are likely not very useful for the problem. Finally, the PyTorch GPU implementation could be extended to the MINIROCKET model.